\documentclass[10pt,twocolumn,letterpaper]{article}

\usepackage{cvpr}
\usepackage{times}
\usepackage{epsfig}
\usepackage{graphicx}
\usepackage{amsmath}
\usepackage{amssymb}
\usepackage[T1]{fontenc}
\usepackage{bm}
\usepackage{bm}
\usepackage{array}

\usepackage[pagebackref=true,breaklinks=true,colorlinks,bookmarks=false]{hyperref}

\cvprfinalcopy % *** Uncomment this line for the final submission

\def\cvprPaperID{****} % *** Enter the CVPR Paper ID here

\begin{document}
	
	%%%%%%%%% TITLE
	\title{Dual Refinement Feature Pyramid Networks for Object Detection}
	
	\author{Jialiang Ma\\
		Chengdu Institute of Computer Application, Chinese Academy of Sciences\\
		University of Chinese Academy of Sciences\\
		{\tt\small majialiang18@mails.ucas.ac.cn}
		\and
		Bin Chen\\
		University of Chinese Academy of Sciences\\
		IRIAI, Harbin Institute of Technology\\
		{\tt\small chenbin2020@hit.edu.cn}
	}
	
	\maketitle

%%%%%%%%% ABSTRACT
\begin{abstract}
	FPN is a common component used in object detectors, it supplements multi-scale information by adjacent level features interpolation and summation. However, due to the existence of nonlinear operations and the convolutional layers with different output dimensions, the relationship between different levels is much more complex, the pixel-wise summation is not an efficient approach. In this paper, we first analyze the design defects from pixel level and feature map level. Then, we design a novel parameter-free feature pyramid networks named Dual Refinement Feature Pyramid Networks (DRFPN) for the problems. Specifically, DRFPN consists of two modules: Spatial Refinement Block (SRB) and Channel Refinement Block (CRB). SRB learns the location and content of sampling points based on contextual information between adjacent levels. CRB learns an adaptive channel merging method based on attention mechanism. Our proposed DRFPN can be easily plugged into existing FPN-based models. Without bells and whistles, for two-stage detectors, our model outperforms different FPN-based counterparts by 1.6 to 2.2 AP on the COCO detection benchmark, and 1.5 to 1.9 AP on the COCO segmentation benchmark. For one-stage detectors, DRFPN improves anchor-based RetinaNet by 1.9 AP and anchor-free FCOS by 1.3 AP when using ResNet50 as backbone. Extensive experiments verifies the robustness and generalization ability of DRFPN. The code will be made publicly available.
	
\end{abstract}

%%%%%%%%% BODY TEXT

\section{Introduction}

\begin{figure}[t]
	\begin{center}
		%\includegraphics[width=0.8\linewidth]{egfigure.eps}
		\includegraphics[width=0.9\linewidth]{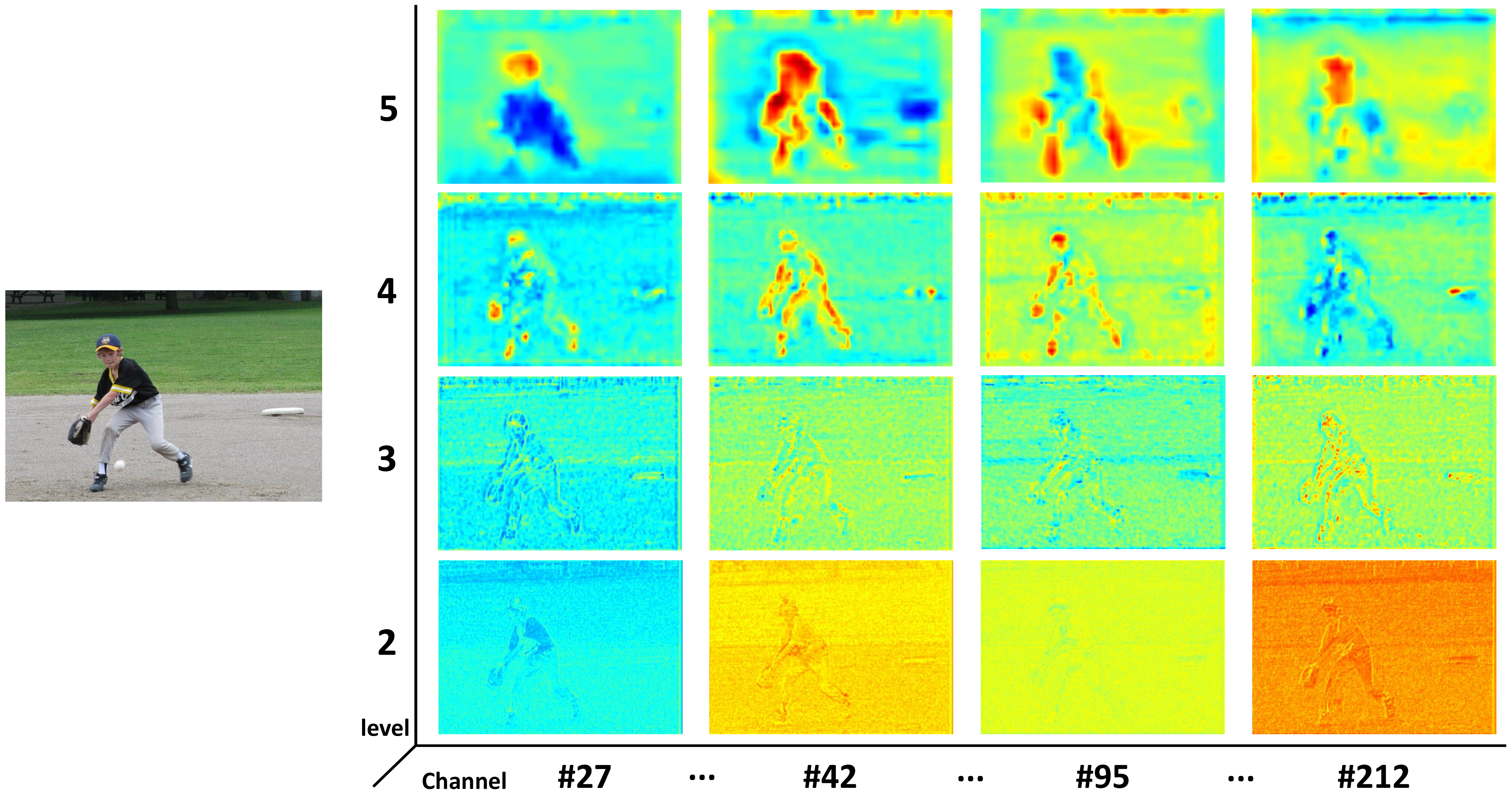}
	\end{center}
	\caption{Visualization of several feature maps from adjacent levels, where the maps are from $27^{th}, 42^{nd}, 95^{th}, 212^{th}$ channels, respectively. All feature maps are from Faster R-CNN~\cite{ren2015faster} using ResNet50~\cite{he2016deep} as backbone. Larger values are denoted by hot colors and vice versa. }
	\label{fig:Fig1}
\end{figure}

\begin{figure*}[t]
	\begin{center}
		%\includegraphics[width=0.8\linewidth]{egfigure.eps}
		\includegraphics[width=0.8\linewidth]{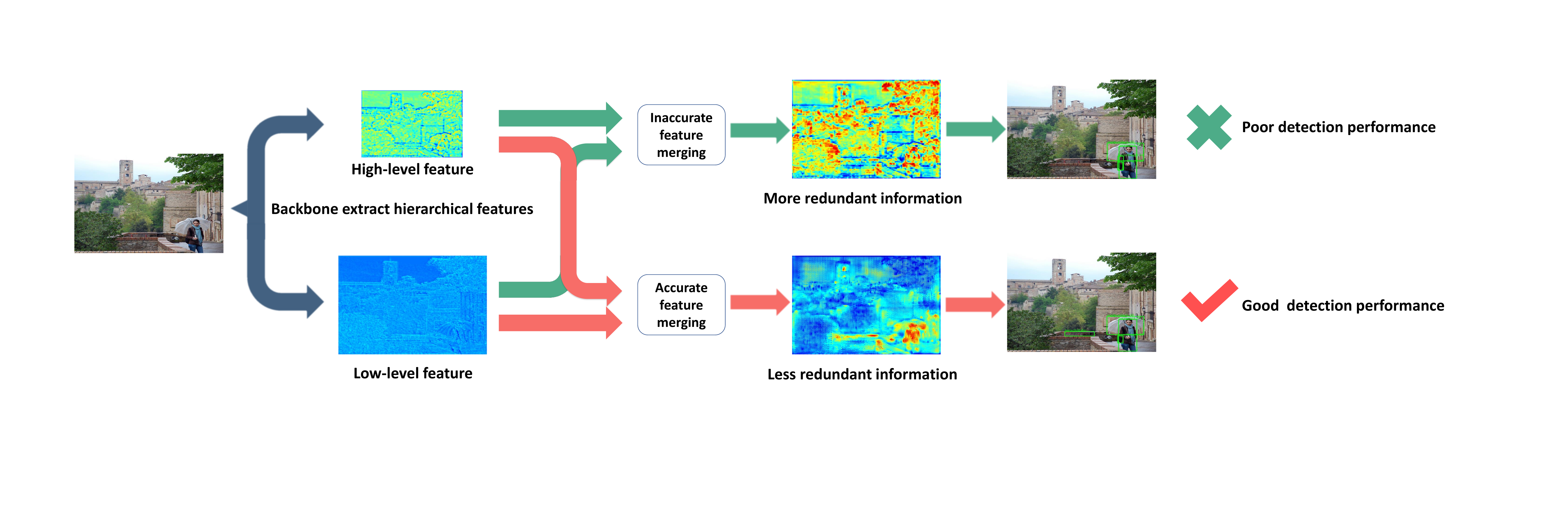}
	\end{center}
	\caption{Inaccurate feature merging methods cause the problems of information redundancy and boundary confusion, which leads to poor detection performance. In this paper, we propose a novel method to improve this situation effectively.
	}
	\label{fig:Fig2}
\end{figure*}

%Object detection consists of two tasks. One is recognition, distinguishing foreground objects from background and assigning them the proper object class labels. The other is the localization, assigning accurate bounding boxes to different objects. 
With the development of convolutional neural networks, object detection has achieved remarkable progress. A series of one-stage detectors~\cite{liu2016ssd:,redmon2016you,lin2017focal,tian2019fcos:,duan2019centernet} and two-stage detectors~\cite{girshick2015fast, ren2015faster,lin2017feature,he2017mask,liu2018path,cai2018cascade} have been proposed, which steadily promotes the state-of-the-art level on several datasets~\cite{lin2014microsoft,everingham2010the}. Among them, FPN~\cite{lin2017feature} is the most commonly used component, which can effectively overcome the problem of the variable scale during detection. Whether it is a one-stage or two-stage detector, the final results can be improved with the help of FPN. Specifically, FPN first uses 1x1 convolution layers to reduce the channels of features extracted from the backbone. Then it constructs a top-down pathway to propagate the multi-scale features from high levels into low levels. This process consists of two stages: First, upsampling each channel of the coarser-resolution feature maps. Second, merging channels with the same spatial resolution. %This process does not require parameters, which is simple and effective.
So far, FPN-based methods have significantly improved the performance of object detection. However, after visualizing the feature maps from the adjacent layers in Figure.~\ref{fig:Fig1}, we find that each of the above stages has a significant shortcoming:

%-------------------------------------------------------------------------

\textbf{Inaccurate spatial sampling.} The most widely used feature upsampling operators are the nearest neighbor and bilinear interpolations. The interpolation operation is a position linear mapping of the sampling point according to the distance in two directions. However, due to the existence of many convolutional operators with different dimensions and repeated pooling operations, there are semantic and resolution gaps between features from adjacent levels. The interpolation introduces much redundant or even error information \textit{at the pixel level} to the following merging. It is not accurate enough to upsample the feature maps only based on the location information. This is one of the reasons for poor detection performance.

%and poor detection of objects similar to the background.
%Second, the interpolation operation only considers a small sub-pixel neighborhood (nearest neighbor 1$\times$1, bilinear 2$\times$2), and cannot capture the rich semantic information required by dense detection tasks.

\textbf{Inaccurate channel fusion.} FPN uses direct addition to fuse the upsampled channels. But for the corresponding channels of adjacent layers, the patterns contained are also different. Some patterns contain more abstract information that is important for detecting the existence of objects, while others contain more detailed information that is helpful for understanding the boundaries of the object. The importance of these patterns is different when merging. Learning the relationship between different patterns \textit{at the feature map level} is beneficial to accurately distinguish the object from the background.

%\textbf{Lack of contextual information \& Insufficient receptive field.} Objects of different sizes are detected on different levels in FPN. Small objects such as baseballs are detected on low-level layers, but baseballs often appear with large objects such as people (athletes) as demonstrated in Figure~\ref{fig:Fig1}. The lack of global scene context information is harmful to object recognition and detection. What's more, convolution operations in convolutional neural networks only focus on local areas. If the receptive field is not large enough, the extracted feature information is limited, and the detection effect is relatively poor. Especially for large-size objects, it is more likely that the receptive field and the sizes do not match. In most FPN based object detection frameworks, even at the highest level, the receptive field can hardly reach the input size. In this paper, we think that the global scene context itself is a way to expand the receptive field, so we define it as the same problem.

As shown in Figure.~\ref{fig:Fig2}, in the top-down feature merging, due to the problems of inaccurate spatial sampling and inaccurate channel fusion in the original FPN, each merged feature map contains a large amount of redundant information, and the boundary between the object and the surrounding is not clear. This situation leads to two problems in the detection results: First, there are too many false positive results. Second, the objects that are similar to the background are not detected accurately or even undetected. To tackle the problems, the latest works build duplicate pathways with more dense connections~\cite{liu2018path, sun2019deep, ghiasi2019nas-fpn:, tan2019efficientdet:, chen2020FPG}, or use feature maps from all levels during detection~\cite{liu2018path, tan2019efficientdet:, guo2019augfpn:} to strengthen the feature representation ability. However, due to missing the core of the problems, all these works inevitably make the network more complex and the computing burden more heavy.

In this paper, we propose \textbf{D}ual \textbf{R}efinement \textbf{F}eature \textbf{P}yramid \textbf{N}etworks (DRFPN), a simple and parameter-free feature pyramid network that integrates two different modules to deal with the problems above respectively. 
%Without significantly increasing the network parameters, the above-mentioned problems are solved by several improvements on FPN. 
The first one is \textbf{S}patial \textbf{R}efinement \textbf{B}lock (SRB), which adaptively learns the location and content of sampling points according to the information from adjacent layers. The second one is \textbf{C}hannel \textbf{R}efinement \textbf{B}lock (CRB), which is a module based on the attention mechanism to learn the importance of channels during merging.

\par We compare the performance of our proposed DRFPN with FPN in object detection/instance segmentation pipelines. Without bells and whistles, DRFPN has achieved the following absolute gains: 1) For object detection, when using different backbones or different learning schedules, the performance of DRFPN-based Faster R-CNN is 1.7 to 2.2 Average Precision (AP) higher than FPN-based counterparts. DRFPN can also be easily plugged into the one-stage detectors. By replacing FPN with DRFPN, RetinaNet and FCOS increased by 1.9 AP and 1.3 AP respectively. 2) For instance segmentation, when using different backbone networks or cascade strategy, the performance of DRFPN-based Mask R-CNN is 1.5 to 1.9 Average Precision (AP) higher than FPN-based counterparts. The experimental results prove the generality of our proposed method.

We summarize our contributions as follows:
\begin{itemize}
	\setlength{\itemsep}{0pt}
	\item The inaccurate feature merging method in the original FPN prevents the multi-scale features from being fully exploited, we explores the inherent shortcomings of FPN design from spatial sampling and channel fusion.
	\item To solve problems, we design two modules: Spatial Refinement Block and Channel Refinement Block, and propose a new feature pyramid network named DRFPN.
	\item We evaluate the DRFPN equipped with various detectors and backbones on MS COCO. Compared with the FPN-based detectors, it consistently brings significant improvements.
\end{itemize}

\begin{figure*}
	\begin{center}
		%\fbox{\rule{0pt}{2in} \rule{.9\linewidth}{0pt}}
		\includegraphics[width=0.7\linewidth]{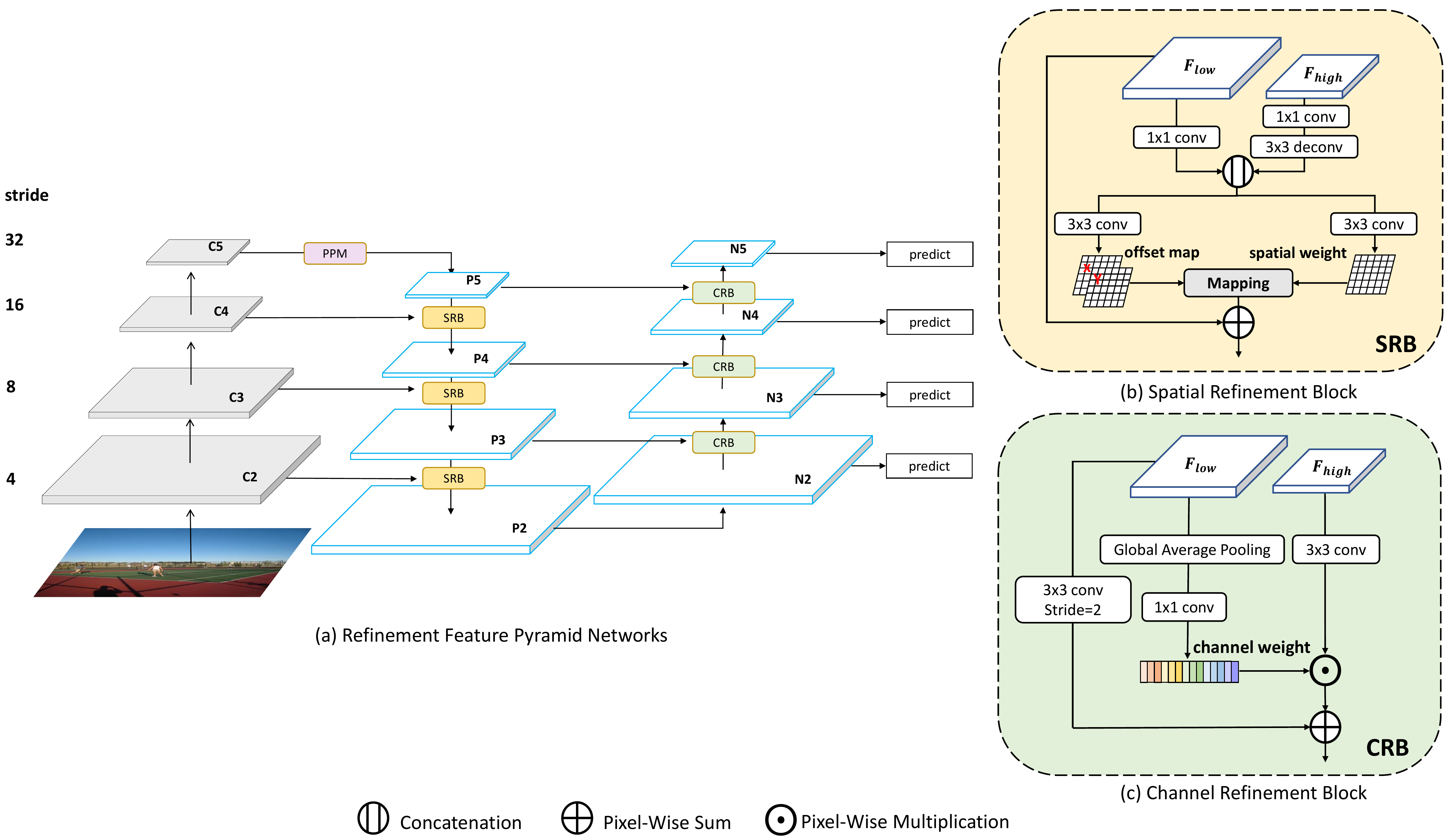}
	\end{center}
	\caption{(a) Overview of the proposed Dual Refinement Feature Pyramid Networks. SRM: Spatial Refinement Block. CRB: Channel Refinement Block. PPM: Pyramid Pooling Module~\cite{zhao2017pyramid}. (b)The details of SRB. (c) The details of CRB.}
	\label{fig:Fig3}
\end{figure*}

\section{Related Work}
\textbf{Object Detectors.} Benefiting from the development of deep learning, object detection has been significantly improved in recent years. Contemporary object detection methods mainly follow two paradigms, two-stage detectors and one-stage detectors. The two-stage detectors~\cite{girshick2014rich, girshick2015fast, ren2015faster} generate a series of bounding boxes by a regional proposal algorithm firstly, then send all the boxes to the convolutional neural networks for classification and regression. Mask R-CNN~\cite{he2017mask} proposes RoI Align to solve the quantization problem of RoI pooling. Cascade R-CNN~\cite{cai2018cascade} notices the difference of RoI hypothesis distribution between the training and inference stages, and introduces a multi-stage refinement network. Recently, a variety of manually designed~\cite{liu2018path, sun2019deep,chen2020FPG} and NAS-based~\cite{xu2019auto-fpn:, ghiasi2019nas-fpn:} pyramid networks further improve the feature representation capability of the detection framework. 

Compared with two-stage detectors, one-stage detectors have a slightly worse performance, but has a faster speed. YOLO~\cite{redmon2018yolov3:} directly predicts the positions and categories of the bounding boxes in an end-to-end manner. SSD~\cite{liu2016ssd:} improves the detection performance by detecting bounding boxes on multiple size feature maps. RetinaNet~\cite{lin2017focal} proposes focal loss to solve the imbalance between foreground and background samples. FCOS~\cite{tian2019fcos:} is an anchor-free object detector, which introduces Center-ness to further improve the detection performance. YOLACT~\cite{bolya2019yolact:} decomposes the instance segmentation task into two simpler parallel tasks, one predicts the prototype mask, the other predicts the mask coefficients. Other object detection algorithms~\cite{law2018cornernet:, duan2019centernet:, zhou2019bottom-up} represent the object with one or a few points, which ensures a balance between speed and accuracy.

\textbf{Feature Modeling.} %Modeling of Spatial Transformations    Guided Anchoring~\cite{wang2019region} and CentripetalNet~\cite{dong2020CentripetalNet} uses semantic features to guide anchors generation and corner points generation.  
There are mainly two paradigms related to feature modeling. The first paradigm is spatial transformation modeling, and the other paradigm is to build a context-sensitive sampling window based on the attention mechanism. Follow the first paradigm, STN~\cite{jaderberg2015spatial} normalizes the entire image by learning global and sparse spatial transformation features (e.g., rotation and scaling). DCN~\cite{dai2017deformable} learns the sampling range of the convolution kernel to adaptively expand the receptive field. Deformable Kernels~\cite{gao2020deformable} deforms the data in kernel space towards specific modality while the receptive field remains unchanged. In video processing and scene parsing tasks, FlowNet~\cite{ilg2017flownet} uses information between adjacent frames to estimate optical flow, DBES~\cite{Li2020DBES} and SFNet~\cite{Li2020SFNet} model the deformation of the entire feature map according to multi-scale information. %These works also focus on the pixel offset in interpolation sampling, but essentially they only reorganize the pixels of the upsampled feature map, and still cannot make full use of global information. 
Follow the second paradigm, CARAFE~\cite{wang2019carafe:} sets a larger sliding window, and uses context information in the sliding window to learn the upsampling. LIP~\cite{gao2019lip:} learns the importance of sampling points by extracting an attention-based map, and then weighting the original feature map to learn the downsampling. Different from previous works, we learn a content-aware upsampling method based on feature maps from adjacent levels to reduce redundant information after feature merging.

\textbf{Channel Attention.} SENet~\cite{hu2019squeeze-and-excitation} uses the channel attention mechanism to allow the model to automatically learn the importance of different channels. But our work focuses on the channel correlation between features from different levels. AugFPN\cite{guo2019augfpn:} uses the attention mechanism to learn the selection of features from different levels during ROI pooling. FPT~\cite{zhang2020FPT} uses channel self-attention mechanism to render finer-resolution feature maps to coarser-resolution feature maps. Different from them, we care about modeling the channel relationship based on the content of adjacent layers in the process of information propagation. In the field of semantic segmentation, there is also a lot of works focusing on channel attention mechanisms~\cite{zhang2018context, li2018pyramid, fu2019dual}. In addition, SCA-CNN~\cite{chen2017sca-cnn:} applies the channel attention mechanism in the field of image captioning. Compared with these works, we focus on learning the weights of feature maps from adjacent levels to guide feature merging.

\textbf{Contextual Information.} Many object detection frameworks prove the importance of context information. AugFPN~\cite{guo2019augfpn:} adds a ratio-invariant supplement composed of adaptive pooling at the highest level of FPN. ACFPN~\cite{Cao2020ACFPN} adds DenseASPP~\cite{yang2018denseaspp} to the highest level of FPN to increase the model receptive field. ThunderNet~\cite{qin2019thundernet:} merges multi-scale information on the top-level features to expand the receptive field. In the field of scene parsing, adaptive pooling layer~\cite{zhao2017pyramid} and atrous convolution~\cite{chen2018deeplab:} are the main methods to extract context. Supplementing contextual information is not the main focus of our work, so we tried several previous works~\cite{zhao2017pyramid, wang2018non-local, yang2018denseaspp} and embedded one of them into our method.

\section{Methodology}

The overall framework of DRFPN is shown in Figure.~\ref{fig:Fig3}(a). To tackle the issues mentioned in the first section, we design two modules: Spatial Refinement Block (SRB) and Channel Refinement Block (CRB).
% Mathematically,
%\begin{equation}
%\begin{split}
%	\tilde{F}_{B} &= F_{CRB}(F_{SRB})\\
%	N_i = &\sum_{l \in Pre(i)}{\tilde{F}_{B}(C_l,C_{l-1})}.
%\end{split}
%\end{equation}
In this section, we will introduce our proposed components first, then describe the entire network architecture. Following the setting of FPN, the hierarchical features used to build the feature pyramid are denoted as ~\bm{$\{C_2, C_3, C_4, C_5\}$}, which the corresponding strides are $\{4, 8, 16, 32\}$ pixels. We define the outputs produced by our proposed DRFPN as ~\bm{$\{N_2, N_3, N_4, N_5\}$}.

\subsection{Spatial Refinement Block}

\begin{figure}[t]
	\begin{center}
		%\includegraphics[width=0.8\linewidth]{egfigure.eps}
		\includegraphics[width=0.8\linewidth]{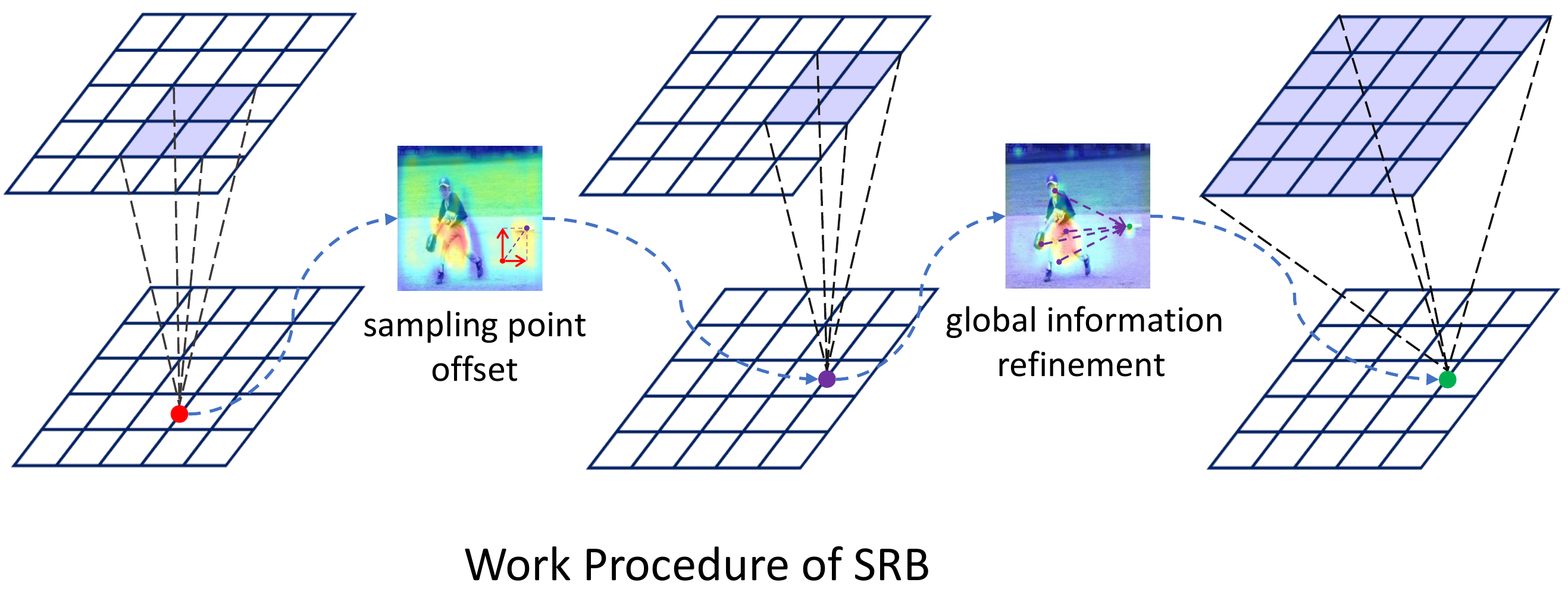}
	\end{center}
	\caption{The total work procedure of SRB.}
	\label{fig:Fig4}
\end{figure}

%As introduced in the first section, the information propagation between adjacent layers in FPN uses bilinear interpolation. However, there are two obvious defects: First, a large number of convolutional layers and pooling layers make the sampling points of the interpolation mapping shift, only using position information to align the resolutions of adjacent layers is a sub-optimal choice. Second, the receptive field of interpolation operation is too small. Inspired by Deformable Convolutional Networks~\cite{dai2017deformable} in the object detection task and SFNet~\cite{Li2020SFNet} in the scene parsing task, we propose a new context-sensitive Spatial Refinement Block (SRB).
The Spatial Refinement Block mainly solves two problems: First, upsampling the coarser-resolution feature maps based only on location information is inaccurate. Second, different sampling points contain different semantic information and should not have the same weight. Inspired by deformable convolution~\cite{dai2017deformable} and spatial attention~\cite{chen2017sca-cnn:}, we first merge the context information of adjacent layers to learn the offsets of sampling points, and then use global information to refine the value of each sampling point. As shown in Figure.~\ref{fig:Fig4}, the overall work procedure of SRB includes two subtasks: sampling point offset and global information refinement.

%In order to solve the problem, we propose a novel context-aware Spatial Refinement Block (SRB), which uses the context information of adjacent layers to learn the sampling point location.

As shown in Figure.~\ref{fig:Fig3}(b), given two adjacent feature maps ~\bm{$F_{l}$} and~\bm{$F_{l-1}$}, Next, we first use two 1$\times$1 convolutional layers to compress their channels to reduce the computational cost, and upsample ~\bm{$F_{l}$} to the same size as ~\bm{$F_{l-1}$} by a deconvolution layer. Then, we concatenate them together, and use the concatenated feature map as the input of a subnet that contains two convolutional layers with the kernel size of 3x3. In order to facilitate the model convergence, we use the offsets of the coordinates to indicate the location of the sampling points, and use the rearranged weight of the sampling points to merge the context information. The subnet contains two outputs: one is the offset map of sampling points in two directions $\delta \in \mathbb{R} ^{2 \times H_{l-1} \times W_{l-1}}$, the other is the rearranged weight $\omega \in \mathbb{R}^{1 \times H_{l-1} \times W_{l-1}}$of each pixel. Mathematically,
\begin{equation}
\begin{split}
\delta &= conv_{1} (cat(deconv(\bm{F_{l}}) , \bm{F_{l-1}})) \\
\omega &= conv_{2}(cat(deconv(\bm{F_{l}}) , \bm{F_{l-1}})),
\end{split}
\end{equation}
where $cat(\cdot)$ means concatenate operation, $conv_1(\cdot)$ is the 3x3 convolutional layer with 2 channels, $conv_2 (\cdot)$ is the 3x3 convolutional layer with 1 channel, $deconv$ is the 3x3 deconvolution layer.
We define the position of the upsampled feature map $p_{l}$ is sampled from the position of the coarser-resolution map $p_{l-1}$, then we add the offset $\delta(p_{l-1})$ to $p_{l-1}$. For training stability, we divide the offset $\delta$ by the mean value of the length and width of ~\bm{$F_{l-1}$}. Mathematically, there is the following mapping relationship,
\begin{equation}
p_{l} \leftarrow p_{l-1}+ \frac{2 \delta (p_{l-1})}{(H_{l-1}+W_{l-1})}.
\end{equation}
In order to solve the quantization problem caused by the floating point offset, we use the differentiable bilinear sampling mechanism proposed in the spatial transformation network~\cite{jaderberg2015spatial}, which takes the four pixel values of the nearest neighbors of $p_{l}$ to approximate the output. After all pixels are mapped, they form a feature map with the same size as the lower-level feature map, denoted as ~\bm{$\widetilde{F}_{l-1}$}. Each position of ~\bm{$\widetilde{F}_{l-1}$} is calculated as follows:
\begin{equation}
~\bm{\widetilde{F}_{l-1}}(p_{l-1}) = \bm{F_{l}}(p_l) = \sum_{p \in \mathcal{N}(p_{l})} w_p \bm{F}_l(p),
\end{equation}
where $\mathcal{N}(p_{l})$ represents the neighbors of pixel $p_l$ in $\bm{F}_l$, and $w_p$ represents the bilinear kernel weights estimated by the distance.

Until now, we have successfully used the contextual information between adjacent layers to refine the location of the sampling points. Next, we use the global information $\omega$ to further refine each pixel of the generated feature map ~\bm{$\widetilde{F}_{l-1}$}. Specifically, we directly multiply the ~\bm{$\widetilde{F}_{l-1}$} by weights $\omega$. Finally, we add the result to the low-level feature map, and then use a 3x3 convolutional layer to get the SRB output $\bm{P}_{l-1}$, Mathematically,
\begin{equation}
\bm{P_{l-1}} = conv_3((\omega \odot \bm{\widetilde{F}_{l-1}} +  up(\bm{{F}_{l}}))+\bm{F_{l-1}}),
\end{equation}
where $conv_3(\cdot)$ is the 3x3 convolutional layer, $up(\cdot)$ means the upsampling operation.

There are several significant differences between our work and DCN~\cite{dai2017deformable}. First, DCN learns offsets for the convolution kernels at each position, while SRB learns offsets for each pixel of the entire feature map. The amount of parameters of the offset matrix has been reduced a lot. Second, SRB combines information of different scales to accurately determine the sampling locations. Finally, the focus of SRB is not only sampling point offset, but also using global information to refine sampling weights, which makes sampling results more accurate.

\subsection{Channel Refinement Block}
%As mentioned in the first section, direct merging without considering the semantic correlation between channels will weaken the representation ability of the feature map. 
Inspired by SENet~\cite{hu2019squeeze-and-excitation}, we proposes Channel Refinement Block (CRB), which aims at optimizing the fusion method between channels according to the context. As shown in Figure.~\ref{fig:Fig3}(c), this module works in a bottom-up fashion, using the channel attention mechanism to guide the network to learn the weight of the channels when adjacent layers are merged.

In our implementation, given two adjacent feature maps ~\bm{$P_{l}$} and ~\bm{$P_{l+1}$}, we first obtain the weight-guided information $\alpha \in \mathbb{R} ^{C \times 1 \times 1}$ of each channel from the low-level feature ~\bm{$P_{l}$} by a global average pooling layer. Then, $\alpha$ goes through a 1$\times$1 convolution to increase the mapping ability. Mathematically,
\begin{equation}
\alpha = conv_4(GAP(\bm{P_{l}})),
\end{equation}
\noindent
where $conv_4$ is a convolutional layer with a kernel size of 1$\times$1, and $GAP$ represents the global average pooling layer. We use a 1$\times$1 convolution to refine the high-level feature ~\bm{$P_{l+1}$}, and combine refined ~\bm{$P_{l+1}$} with the weight-guided information $\alpha$ by Hadamard product. Next, we use a 3x3 convolution with a step size of 2 to reduce the scale of low-level features ~\bm{$P_{l}$}, and add the corresponding elements of  ~\bm{$P_{l}$} and ~\bm{$P_{l+1}$}. Lastly, The fused feature map is processed by another 3x3 convolutional layer to generate ~\bm{$N_{l+1}$} for following sub-networks. Above steps can be calculated as follows:
\begin{equation}
\begin{split}
\bm{P_{l+1}} = conv_5&(\bm{P_{l+1}})  \odot \alpha + conv_{down}(\bm{P_{l}}) \\
\bm{N_{l+1}} &= conv_6(\bm{P_{l+1}}),
\end{split}
\end{equation}
\noindent
where $conv_5(\cdot)$ is a 3$\times$3 convolution, $conv_{down}(\cdot)$ is a 3x3 stride convolution, and $conv_6(\cdot)$ is a 1$\times$1 convolution.

%It is worthy to note that CRB is embedded in the bottom-up pathway extended after FPN, instead of the top-down pathway, which is the key reason why this module works. As shown in Figure.~\ref{fig:Fig2}(a), the captured patterns of adjacent channels are not consistent. Since global average pooling is a parameter-free operation, there is a gap between the channel features extracted by global average pooling and the low-level feature map. Direct multiplication will weaken the low-level feature representation. We use the hierarchical features constructed by the top-down pathway as the input of CRB. As shown in Figure.~\ref{fig:Fig2}(b), the output of the top-down pathway merges multi-scale features while roughly aligning the patterns captured by adjacent layers. In addition, considering that global average pooling is to directly average the pixels of the feature map, it needs rich pixel information. We use the finer-resolution feature map to guide the learning of channel relationships. Without increasing calculation significantly, CRB captures more accurate semantic relationship between channels and improves the final detection performance. In the ablation experiment, we prove that it is not feasible to directly apply the global average pooling operation in the top-down pathway.
Different from the SE Block~\cite{hu2019squeeze-and-excitation}, our work focuses on how adjacent layers guide each other by context when merging, and uses semantic information from low-level feature maps to learn the weights of high-level feature maps during feature fusion. %Considering that global average pooling is to directly average the pixels of the feature map, it needs rich pixel information. We use the finer-resolution feature map to guide the learning of channel relationships.
Without increasing calculation significantly, CRB captures more accurate semantic relationship between channels and improves the final detection performance.

%CRB is different from the previous work~\cite{hu2019squeeze-and-excitation}.The SENet block~\cite{hu2019squeeze-and-excitation} uses a global average pooling layer and fully connected layers to learn the weights of channels, which increases a lot of computation. But our work focuses on how adjacent layers guide each other by context when merging. %SCA-CNN~\cite{chen2017sca-cnn:} applies the channel attention mechanism in the field of image captioning, but our work is aimed at finding the relationship between channels in adjacent layers. At the same time, because of the softmax activation function in SCA-CNN, the model is time-consuming and too difficult to capture the negative weight between channels. 

%------------------------------------------------------------------------
\begin{table*}
	\begin{center}
		%\tiny
		\scriptsize
		\begin{tabular}{c c p{0.8cm}<{\centering} c c c c c c}
			\hline
			Method & Backbone & Schedule & $AP_{box}$ & $AP_{50}$ & $AP_{75}$ & $AP_{S}$ & $AP_{M}$ & $AP_{L}$\\
			\hline
			%RetinaNet~\cite{lin2017focal} & ResNet-101 & - & 39.1 & 59.1 & 42.3 & 21.8 & 42.7 & 50.2 \\
			%CornetNet~\cite{law2018cornernet:} & Hourglass-104 & - & 40.5 & 56.5 & 43.1 & 19.4 & 42.7 & 53.9 \\
			%FCOS~\cite{tian2019fcos:} & ResNet-101 & - & 41.5 & 60.7 & 45.0 & 24.4 & 44.8 & 51.6 \\
			%Faster R-CNN~\cite{ren2015faster} & ResNet-101 & - & 36.2 & 59.1 & 42.3 & 21.8 & 41.7 & 50.2 \\
			%Mask RCNN~\cite{he2017mask} &ResNet-101 & - &38.2 &60.3 &41.7 &20.1 &41.1 &50.2 \\
			%Cascade R-CNN~\cite{cai2018cascade}&ResNet-101 & - & 42.8 &62.1 &46.3 &23.7 &45.5 &55.2 \\
			%\hline
			Libra R-CNN~\cite{pang2019libra} & ResNet-50 & 1$\times$ & 38.7 & 59.9 & 42.0 & 22.5 & 41.1 & 48.7 \\ 
			Libra R-CNN~\cite{pang2019libra} & ResNet-101 & 1$\times$ & 40.3 & 61.3 & 43.9 & 22.9 & 43.1 & 51.0 \\ 
			Faster R-CNN w/ LIP~\cite{gao2019lip:} & ResNet-50 & 2$\times$ & 39.2 & 61.2 & 42.5 & 24.0 & 43.1 & 50.3 \\
			Faster R-CNN w/ CARAFE~\cite{wang2019carafe:} & ResNet-50 & - & 38.1 & 60.7 & 41.0 & 22.8 & 41.2 & 46.9 \\
			%Faster R-CNN w/ LIP~\cite{gao2019lip:} & ResNet-50 & - & 39.2 & 61.2 & 42.5 & 24.0 & 43.1 & 50.3 \\
			Faster R-CNN w/ AugFPN~\cite{guo2019augfpn:}& ResNet-50 & 1$\times$ & 38.8 & 61.5 & 42.0& 23.3 & 42.1 & 47.7 \\
			Faster R-CNN w/ PConv~\cite{wang2020SEPC}& ResNet-50 & 1$\times$ & 38.5 & 59.9 & 41.4& - & - & - \\
			%Faster R-CNN w/FPG 9@256~\cite{chen2020FPG}& ResNet-50& 1$\times$ & 39.2 & 60.8 & 42.7 & 22.7 & 41.9 & 48.4 \\
			%Faster R-CNN w/FPG 9@256~\cite{chen2020FPG}& ResNet-101& 1$\times$ & 40.6 & 62.2 & 44.3 & 23.4 & 43.5 & 50.6 \\
			NAS-FPN 7@256~\cite{chen2020FPG, ghiasi2019nas-fpn:}& ResNet-50& 1$\times$ & 39.0 & 59.5 & 42.4 & 22.4 & 42.6 & 47.8 \\
			NAS-FPN 7@256~\cite{chen2020FPG, ghiasi2019nas-fpn:}& ResNet-101& 1$\times$ & 40.3 & 61.2 & 43.8 & 23.1 & 43.9 & 50.1 \\
			Mask R-CNN w/ FPT~\cite{zhang2020FPT} & ResNet-101 & 2$\times$ & 41.6 & 60.9 & 44.0 & 23.4 & 41.5 & 53.1 \\
			\hline
			RetinaNet* & ResNet-50 & 1$\times$ & 36.1 & 54.8 & 38.1 & 19.1 & 39.7 & 47.0 \\
			FCOS* & ResNet-50 & 1$\times$ & 36.6 & 55.7 & 38.8 & 20.7 & 40.1 & 47.4 \\
			Faster R-CNN* & ResNet-50 & 1$\times$ & 37.4 & 58.1 & 40.4 & 21.2 & 41.0 & 48.1 \\
			Faster R-CNN* & ResNet-101 & 1$\times$ & 39.4 & 60.1 & 43.1 & 22.4 & 43.7 & 50.0 \\
			Faster R-CNN* & ResNet-101 & 2$\times$ & 39.8 & 60.1 & 43.3 & 22.5 & 43.6 & 51.8 \\
			Faster R-CNN* & ResNeXt-64x4d-101 & 1$\times$ & 41.7 & 64.1 & 45.3 & 25.1 & 45.1 & 52.1 \\
			Mask R-CNN* & ResNet-50 & 1$\times$ & 38.2(34.4) & 58.8(55.7) & 41.4(37.2) & 21.9(18.3) & 40.9(37.4) & 49.1(46.1) \\
			Mask R-CNN* & ResNet-101 & 1$\times$ & 40.0(36.1) & 60.5(57.5) & 44.0(38.6) & 22.6(18.8) & 44.0(39.7) & 51.9(48.9) \\
			Mask R-CNN* & ResNet-101 & 2$\times$ & 40.8(36.6) & 61.0(57.9) & 44.5(39.1) & 23.0(19.2) & 45.0(40.2) & 53.5(49.2) \\
			Cascade Mask R-CNN*  & ResNet-50 & 1$\times$ & 41.2(35.9) & 59.4(56.6) & 45.0(38.4) & 23.9(19.4) & 44.2(38.5) & 53.8(48.5) \\
			Cascade Mask R-CNN*  & ResNet-101 & 1$\times$ & 42.9(37.3) & 61.0(58.2) & 46.6(40.1) & 24.4(19.7) & 46.5(40.6) & 56.0(50.9) \\
			\hline
			RetinaNet w/ DRFPN & ResNet-50 & 1$\times$ & 38.0 [\textbf{+1.9}] & 58.1 & 40.6 & 22.2 & 41.5 & 45.6 \\
			%RetinaNet w/ RFPN & ResNet-101 & 1$\times$ &  &  &  &  &  &  \\
			FCOS w/ DRFPN & ResNet-50 & 1$\times$ & 37.9[\textbf{+1.3}] & 57.6 & 40.5 & 21.0 & 41.0 & 47.3 \\
			%FCOS w/ RFPN & ResNet-101 & 1$\times$ &  &  &  &  &   &   \\
			Faster R-CNN w/ DRFPN & ResNet-50 & 1$\times$ & 39.6 [\textbf{+2.2}] & 61.2 & 42.8 & 23.1 & 42.6 & 49.4 \\
			Faster R-CNN w/ DRFPN & ResNet-101 & 1$\times$ & 41.1 [\textbf{+1.7}] & 62.8 & 44.7 & 23.9 & 44.3 & 52.2 \\
			Faster R-CNN w/ DRFPN & ResNet-101 & 2$\times$ & 41.8 [\textbf{+2.0}] & 63.0 & 45.7 & 23.1 & 44.7 & 53.1 \\
			Faster R-CNN w/ DRFPN & ResNeXt-64x4d-101 & 1$\times$ & 43.4 [\textbf{+1.7}] & 65.2 & 47.4 & 25.7 & 47.0 & 54.6 \\
			Mask R-CNN w/ DRFPN & ResNet-50 & 1$\times$ & 40.0 [\textbf{+1.8}](36.3 [\textbf{+1.9}]) & 61.4(58.4) & 43.6(38.9) & 23.2(19.6) & 42.8(38.7) & 50.0(47.2) \\
			Mask R-CNN w/ DRFPN& ResNet-101 & 1$\times$ & 41.9 [\textbf{+1.9}](37.8 [\textbf{+1.7}]) & 63.3(60.2) & 45.7(40.5) & 24.4(20.6) & 45.1(40.5) & 53.0(49.9) \\
			Mask R-CNN w/ DRFPN& ResNet-101 & 2$\times$ & 42.7 [\textbf{+1.9}](38.4 [\textbf{+1.8}]) & 62.9(60.4) & 47.1(41.3) & 23.8(21.0) & 46.5(41.4) & 56.7(51.3) \\
			Cascade Mask R-CNN w/ DRFPN & ResNet-50 & 1$\times$ & 43.0 [\textbf{+1.8}](37.7 [\textbf{+1.8}]) & 61.7(59.2) & 46.6(40.6) & 24.8(20.4) & 45.7(40.0) & 54.5(49.4) \\
			Cascade Mask R-CNN w/ DRFPN & ResNet-101 & 1$\times$ & 44.5 [\textbf{+1.6}](38.8 [\textbf{+1.5}]) & 63.2(60.5) & 48.3(41.9) & 25.6(21.0) & 47.4(41.4) & 57.1(51.6) \\
			\hline
		\end{tabular}
	\end{center}
	\caption{Comparison with the state-of-the-art object detection methods on COCO test-dev. The symbol '*' means our re-implementation results. The number in [] stands for the relative improvement. For the detectors with mask head, the results in () means the corresponding mask results.}
	\label{Table1}
\end{table*}

\subsection{Overall Architecture} 

As described in this section, we design SRB and CRB to optimize different problems. In order to reduce the coupling between two modules during joint optimization, we alternately optimize spatial sampling and channel fusion. Specifically, we simply extend a bottom-up pathway after the lowest level of FPN. Then we embed SRB into the top-down pathway and CRB into the bottom-up pathway.

%Mathematically, the output becomes 
%\begin{equation}
%\begin{split}
%P_i &= \sum_{l \in Pre(i)} F_{SRB} (C_l, C_{l-1}) \\
%N_i &= \sum_{l\in Pre^*(i)} F_{CRB}(P_l, P_{l-1}),
%\end{split}
%\end{equation}
%where $Pre(i)$ means the layers before the i-th layer in the top-down pathway, $Pre^*(i)$ means the layers before the i-th layer in the bottom pathway. 

We additionally adopt the Pyramid Pooling Module ~\cite{zhao2017pyramid} because of its superior power to capture contextual information. In our realization, the output of PPM shares the same resolution as that of the last FPN level. In this situation, we concatenate PPM and the highest-level features extracted by backbone together as the input of the top-down pathway. Other context modules like Non-local~\cite{wang2018non-local} and DenseASPP~\cite{yang2018denseaspp} can also be plugged into our network, which are also experimentally ablated in this paper.

%Different from previous work ~\cite{guo2019augfpn:, liu2018path}, our method does not need to fuse ROI features from different pyramid levels, \emph{which means that our output is exactly the same as that defined in FPN, so it can be used in any FPN-based method}. Moreover, our method does not need to introduce additional loss (e.g., consistent supervision~\cite{guo2019augfpn:}) nor training skills (e.g., DropBlock~\cite{ghiasi2018dropblock:, zhang2020FPT}), which trains quickly and saves computing resources. 

\section{Experiments}

%-------------------------------------------------------------------------
\subsection{Dataset and Evaluation Metrics}
All our experiments use the MS COCO detection dataset containing 80 categories totally. The dataset contains 115k images for training (\emph{train2017}), 5k images for validation (\emph{val2017}), and 20k images for testing (\emph{test-dev}). The labels of ~\emph{test-dev} are not released publicly. We use \emph{train2017} to train the models, use \emph{val2017} for validation, and report ablation study on \emph{val2017}. The final results are reported on \emph{test-dev}. All reported results follow standard COCO-style Average Precision (AP) metrics.	
%-------------------------------------------------------------------------

\subsection{Implementation Details}

All our experiments are based on mmdetection~\cite{chen2019mmdetection:}. Since most baseline methods in this repository perform better than the original papers, we reimplement all baseline methods using mmdetetion for a fair comparison. The input images are resized to have a shorter size of 800 pixels. We use 3 NVIDIA GeForce RTX TITAN GPUs with 24GB memory, set each GPU batch size as 8. In 1$\times$ schedule, models are trained for 12 epochs, the initial learning rate is set as 0.03 and decreases by a ratio of 0.1 in the 8th and 11th epoch. In 2$\times$ schedule, models are trained for 24 epochs. The initial learning rate is set as 0.03 and decreases by a ratio of 0.1 in the 16th and 22th epoch.
%-------------------------------------------------------------------------

\subsection{Main Results}
In this section, we use the COCO \emph {test-dev} dataset to evaluate the performance of DRFPN on the object detection task and compare it with other state-of-the-art one-stage and two-stage detectors. All results are shown in Table~\ref{Table1}, by replacing FPN with DRFPN, Faster R-CNN using ResNet50 as backbone achieves 39.6 $AP_{box}$, which is 2.2 points higher than Faster R-CNN based on ResNet50-FPN. In addition, with a longer learning schedule or a more powerful backbone network, DRFPN can still bring significant performance improvements. For example, when using 2$\times$ schedule ResNet101 or ResNext101-64x4d as feature extractors, our method improves the performance by 2.0 $AP_{box}$ and 1.7 $AP_{box}$ respectively. 
For one-stage detectors, we validate the performance of DRFPN on two different types of detectors, i.e. anchor-based RetinaNet and anchor-free FCOS. DRFPN improves RetinaNet by 1.9 $AP_{box}$ when using ResNet50 as backbone. Meanwhile, FCOS is boosted to 37.9 $AP_{box}$ from 36.6 $AP_{box}$ when replacing FPN with DRFPN.

In order to evaluate the performance of DRFPN on the instance segmentation task. We also built experiments for Mask R-CNN and Cascade Mask R-CNN. As shown in Table~\ref{Table1}, by replacing FPN with DRFPN, Mask RCNN on ResNet50 improves instance segmentation by 1.9 $AP_{mask}$, Cascade Mask RCNN on ResNet50 improves instance segmentation by 1.8 $AP_{mask}$. When using ResNet101 as the backbone network, Mask RCNN on ResNet101 improves instance segmentation by 1.7 $AP_{mask}$, Cascade Mask RCNN on ResNet101 improves instance segmentation by 1.5 $AP_{mask}$. 

\subsection{Visualization of Results}
\begin{figure*}[t]
	\begin{center}
		%\includegraphics[width=0.8\linewidth]{egfigure.eps}
		\includegraphics[width=0.8\linewidth]{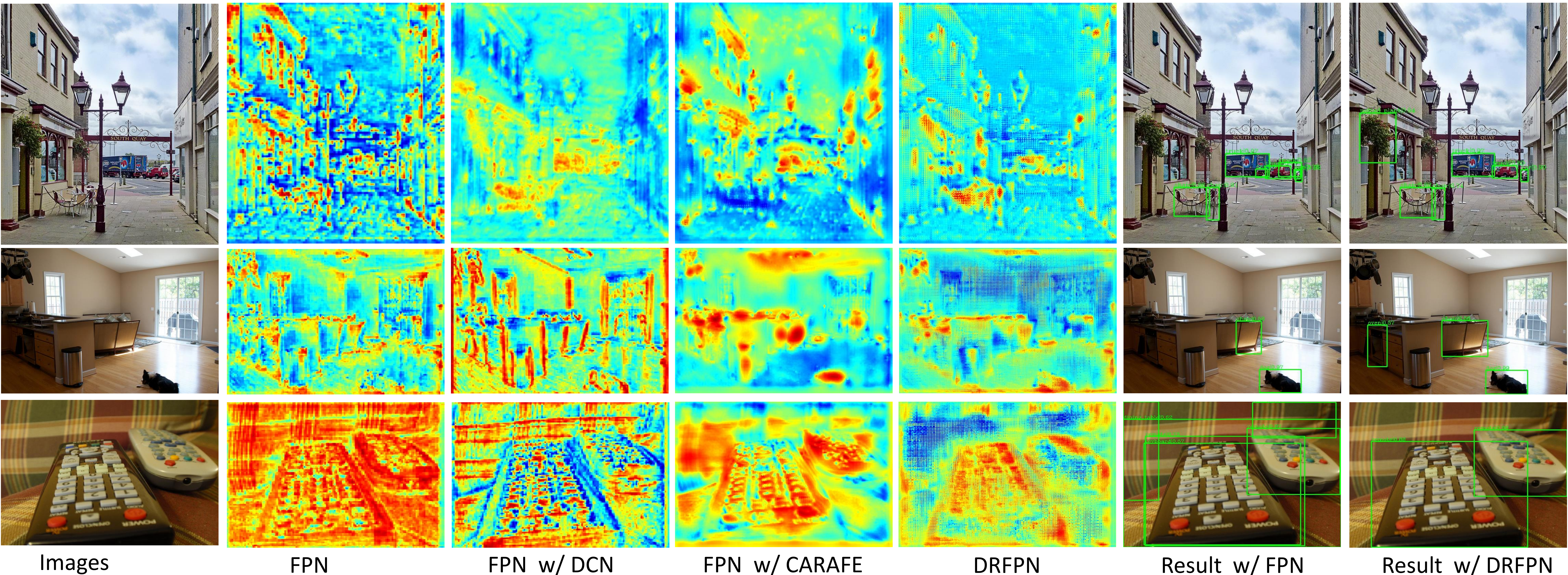}
	\end{center}
	\caption{Visualization of the feature fusion. DRFPN can effectively reduce the information redundancy caused by the fusion of multi-scale features, which makes the final detection results more accurate.}
	\label{fig:Fig5}
\end{figure*}

In this section, we visualize the merging results constructed by DRFPN and several other feature fusion methods~\cite{dai2017deformable, wang2019carafe:}. Specifically, we randomly select a pair of channels from the corresponding position of adjacent layers as input, the outputs of different feature fusion methods are shown in Figure.~\ref{fig:Fig5}. Because the lowest level of the feature pyramid contains the most redundant information generated by the top-down feature merging, we visualize the lowest level features to observe the differences between feature modeling methods. As shown in Figure.~\ref{fig:Fig5}, both DCN~\cite{dai2017deformable} and CARAFE~\cite{wang2019carafe:} focus on the optimization of the kernel, so their effect in the top-down feature merging is more similar to using attention mechanism to increase the saliency of the object areas. Different from them, DRFPN optimizes the entire feature map. The activation areas captured by the model tend to focus on the objects, the features are more structural, and redundant information is significantly reduced. Finally, we obtain the detection results at the threshold = 0.5. The results shows that compared with FPN, DRFPN detects more accurate detection results and fewer false positive boxes. This shows that our SRB and CRB modules effectively reduce the redundant information brought by hard merging, and better capture the feature boundary of the object.

\subsection{Ablation Study}
In this section, we conduct ablation experiments to analyze the performance of each module in our proposed method. The baseline method for all ablation studies is FPN-based Faster R-CNN with ResNet50.

\textbf{Ablation studies for each module.} To analyze the importance of each module in DRFPN, we apply Spatial Refinement Block, Channel Refinement Block and PPM~\cite{zhao2017pyramid} separately to the model to verify the effectiveness. In addition, the improvements brought by combination of different modules are also presented to demonstrate that these modules are complementary to each other. All results are shown in Table~\ref{Table2}.

As shown in Table~\ref{Table2}, when Spatial Refinement Block (SRB), Channel Refinement Block (CRB), Pyramid Pooling Module (PPM) are added separately to FPN, the baseline performance is improved by 0.6, 0.8 and 0.6 AP respectively. The results show that SRB learns a context-aware sampling method based on the information of adjacent layers, which reduces the spatial sampling bias between adjacent layers. CRB uses multi-scale information to learn the channel relationship between adjacent layers, which enhances the representation capability of patterns. The introduction of PPM increases $AP_L$ 1.0 point, which means that for a smaller model like ResNet50, larger receptive field is good for the perception of large objects. It can be seen that when SRB or CRB is introduced simultaneously, the results of objects in small, medium, and large scale are all improved. In addition, the combination of SRB and PPM or CRB and PPM can also improve the baseline by 1.1 and 1.2 AP, which proves that the refined performance of SRB and CRB can also benefit from higher receptive field. When all three components are integrated into the benchmark method, the performance can be improved to 39.1 AP. These results show that these three modules are complementary to each other.

\begin{table}
	\begin{center}
		\scriptsize
		\begin{tabular}{p{5pt} p{5pt} c| c c c c c c }
			\hline
			\textbf{SRB} & \textbf{CRB} & \textbf{PPM} &$AP$ & $AP_{50}$ & $AP_{75}$ & $AP_{S}$ & $AP_{M}$ & $AP_{L}$\\
			\hline\hline
			\multicolumn{3}{c|}{baseline}& 37.2  & 58.1 & 40.0 & 21.2 & 40.9& 48.1\\
			\checkmark & \ & \ & 37.8 & 58.6 & 41.0 & 22.6 & 41.5 & 49.3\\
			\ & \checkmark & \ &  38.0 & 59.7 & 41.0 & 22.6 & 41.8 & 49.4\\
			\ & \ & \checkmark & 37.8 & 59.2 & 41.2 & 22.2 & 41.3& 49.1 \\
			\checkmark & \checkmark & \ & 38.3 & 59.3 & 41.5 & 21.5 & 41.7 & 50.0 \\
			\checkmark & \ & \checkmark & 38.3 & 59.7 & 41.2 & 22.1 & 42.1 & 49.7\\
			\ & \checkmark & \checkmark & 38.4 & 59.6 & 41.5 & 23.1 & 42.2 & 49.4\\
			\checkmark & \checkmark & \checkmark & 39.1 & 60.3 & 42.5 & 22.9 & 43.1 & 50.7 \\
			\hline
		\end{tabular}
	\end{center}
	\caption{Performance of each module. Results are reported on COCO val2017. \textbf{SRB}: Spatial Refinement Block, \textbf{CRB}: Channel Refinement Block, \textbf{PPM}: Pyramid Pooling Module~\cite{zhao2017pyramid}.}
	\label{Table2}
\end{table}	

\begin{table}
	\begin{center}
		\tiny
		\begin{tabular}{l| c c c c c c}
			\hline
			Setting & $AP$ & $AP_{50}$ & $AP_{75}$ & $AP_{S}$ & $AP_{M}$ & $AP_{L}$\\
			\hline\hline
			TD & 37.2  & 58.1 & 40.0 & 21.2 & 40.9& 48.1\\
			TD + BU & 37.3 & 58.6 & 40.8 & 21.4 & 41.0& 48.6\\
			TD + PPM & 37.8 & 59.2 & 41.2 & 22.2 & 41.3& 49.1 \\
			TD + BU + PPM & 38.2 & 59.8 & 41.2 & 22.0 & 41.9& 49.9\\
			\hline
			%TD[SRB+CRB]& 37.2 & 58.1 & 39.8 & 22.0 & 41.1 & 48.5\\ \sdsds
			TD[SRB+CRB]& 37.7 & 59.3 & 40.8 & 21.6 & 41.8 & 48.1\\	
			TD[SRB+CRB]+PPM & 38.1 & 59.4 & 41.4 & 22.4 & 41.8 & 48.6\\
			\hline
			%TD + PPM
			%TD + BU + PPM & 38.2 & 59.8 & 41.2 & 22.0 & 41.9& 49.9\\
			TD[CRB]+BU[SRB] & 38.2 & 59.1 & 41.2 & 22.0 & 41.9 & 50.2 \\	
			TD[SRB]+BU[CRB] & 38.3 & 59.3 & 41.5 & 22.0 & 41.7 & 50.0 \\	
			TD[CRB]+BU[SRB]+PPM & 38.9 & 59.9 & 42.2 & 22.1 & 42.4& 51.4\\	
			TD[SRB]+BU[CRB]+PPM & 39.1 & 60.3 & 42.5 & 22.9 & 43.1& 50.7\\	
			\hline
		\end{tabular}
	\end{center}
	\caption{Ablation performance of network design. TD means top-down pathway, BU means bottom-up pathway. PPM: Pyramid Pooling Module~\cite{zhao2017pyramid}.}
	\label{Table3}
\end{table}

\textbf{Ablation studies for network design.} We study the optimization sequence of SRB and CRB in the feature pyramid pathway. The experimental results are shown in Table~\ref{Table3}. When we simultaneously embed SRB and CRB between adjacent layers in the top-down pathway, the performance is only improved by 0.5 AP. When we embed CRB in the top-down pathway and SRB in the extended bottom-up pathway, we can increase 1.0 AP. When we embed SRB in the top-down pathway and CRB in the extended bottom-up pathway, we can increase 1.1 AP totally, of which $AP_L $ increases by 1.9 points. Obviously, by alternately optimizing spatial sampling and channel fusion, the coupling between two modules and the model optimization difficulty are reduced. It is worthy to note that when we use deconvolution to extend a bottom-up pathway, the performance of the model hardly improves, which seems to be different from the results in the PAFPN~\cite{liu2018path} paper. We consider the reason for this result is that feature fusion based on interpolation or deconvolution is not accurate enough and its effect is limited . As shown in Table~\ref{Table3}, using our proposed SRB and CRB to supplement multi-scale information is 1.0 AP better than just adding a bottom-up pathway. Finally, we append PPM (Pyramid Pooling Module)~\cite{zhao2017pyramid} to capture global contextual information with a larger receptive field. It can be seen that both SRB and CRB can benefit from the global information supplemented by PPM, and the final detection result increases by 1.9 AP.

\begin{table}
	\begin{center}
		\scriptsize
		\begin{tabular}{c c| c c c c c c }
			\hline
			SPO & GIR & $AP$ & $AP_{50}$ & $AP_{75}$ & $AP_{S}$ & $AP_{M}$ & $AP_{L}$\\
			\hline\hline
			\multicolumn{2}{c|}{FPN} & 37.2 & 58.1 & 40.0 & 21.2 & 40.9 & 48.1 \\
			\multicolumn{2}{c|}{FPN w/ DCN} & 38.2 & 59.2 & 41.6 & 22.4 & 41.8 & 48.5 \\
			\multicolumn{2}{c|}{DRFPN w/o SRB} & 38.4 & 59.6 & 41.5 & 23.1 & 42.2 & 49.4 \\
			\hline
			\checkmark & \ & 38.7 & 59.8 & 42.3 & 22.7 & 42.3 & 50.1 \\
			\ & \checkmark & 38.6 & 59.7 & 42.2 & 23.0 & 42.2 & 49.7 \\
			\checkmark & \checkmark & 39.1 & 60.3 & 42.5 & 22.9 & 43.1 & 50.7\\
			\hline
		\end{tabular}
	\end{center}
	\caption{Ablation performance of Spatial Refinement Block. DCN: Deformable Convolutional Networks~\cite{dai2017deformable}. SPO: sampling point offset, GIR: global information refinement.}
	\label{Table4}
\end{table}	

\begin{table}
	\begin{center}
		\scriptsize
		\begin{tabular}{l| c c c c c c }
			\hline
			Setting & $AP$ & $AP_{50}$ & $AP_{75}$ & $AP_{S}$ & $AP_{M}$ & $AP_{L}$\\
			\hline\hline
			FPN & 37.2  & 58.1 & 40.0 & 21.2 &40.9& 48.1\\
			DRFPN w/o SRB & 38.4 & 59.6 & 41.5 & 23.1 & 42.2 & 49.4\\
			\hline
			$F_{l}$ & 38.4 & 59.8 & 41.9 & 22.3 & 42.0& 49.9\\
			$F_{h}$ & 38.3 & 59.8 & 41.5 & 22.3 & 42.7 & 49.4\\
			add $(F_{l},F_{h})$ & 38.9 & 60.3 & 42.4 & 23.1 & 43.0 & 50.1 \\
			cat $(F_{l},F_{h})$ & 39.1 & 60.3 & 42.5 & 22.9 & 43.1 & 50.7 \\
			\hline
		\end{tabular}
	\end{center}
	\caption{Ablation studies of Spatial Refinement Block on COCO val2017. $F_{l}$ means the low-level feature map, $F_{h}$ means the high-level feature map.}
	\label{Table5}
\end{table}

\textbf{Ablation studies for SRB.}  The results of the SRB ablation experiments are shown in Table~\ref{Table4} and Table~\ref{Table5}. We first study the separate performance of two subtasks in Spatial Refinement Block. From Table~\ref{Table4} we observe that learning the position offset of the sampling point according to the context and correcting the pixel value with global information are both beneficial to the final performance. The implementation of sampling point offset improves the baseline by 1.5 AP. When both sampling point offset and global information refinement are implemented, there is an improvement of 1.9 AP relative to the baseline, especially $AP_L$ has an improvement of 2.6 AP, which proves that global information is very helpful for the detection of large objects. Compared with FPN with DCN, our method has 0.9 AP improvement. In addition, our DRFPN with SRB and CRB has 20\% less training time.

We also experiment with several settings to get the offset and global refinement weight. It can be seen from Table~\ref{Table5}, when we use only high-level feature maps or low-level feature maps to refine the sampling, the result can be improved by \textasciitilde1.2 AP. When adjacent layer features are used together, the result can be improved by \textasciitilde1.8 AP. The results show that the multiscale information from adjacent layers is very important to guide the sampling process accurately.

\begin{table}
	\begin{center}
		\scriptsize
		\begin{tabular}{l| c c c c c c }
			\hline
			Setting & $AP$ & $AP_{50}$ & $AP_{75}$ & $AP_{S}$ & $AP_{M}$ & $AP_{L}$\\
			\hline\hline
			FPN & 37.2  & 58.1 & 40.0 & 21.2 &40.9& 48.1\\
			DRFPN w/o CRB &38.3 & 59.7 & 41.2 & 22.1 & 42.1 & 49.7\\
			\hline
			$F_{l}$ & 39.1 & 60.3 & 42.5 & 22.9 & 43.1& 50.7\\
			$F_{h}$ & 38.4 & 59.9 & 41.3 & 22.1 & 42.1 & 49.9\\
			add $(F_{l},F_{h})$ & 38.9 & 60.1 & 42.2 & 22.2 & 42.9 & 50.6 \\
			cat $(F_{l},F_{h})$ & 38.9 & 60.3 & 42.4 & 23.1 & 43.0 & 50.1 \\
			\hline
		\end{tabular}
	\end{center}
	\caption{Ablation studies of Channel Refinement Block on COCO val2017. $F_{l}$ means the low-level feature map, $F_{h}$ means the high-level feature map.}
	\label{Table6}
\end{table}	

%\begin{table}
%	\begin{center}
%		\tiny
%		\begin{tabular}{l| c c c c c c c }
%			\hline
%			Method & $AP$ & $AP_{S}$ & $AP_{M}$ & $AP_{L}$ & Time & Params & Flops(G) \\
%			\hline\hline
%			FPN &37.2 & 21.2 & 40.9& 48.1 & 1.14s & 41.53M  & 207\\
%			DRFPN w/o PPM & 38.3  & 21.5 & 41.7 & 50.0 & 1.37s & 47.79M & 263\\
%			+PPM~\cite{zhao2017pyramid} & 39.1  & 22.9 & 43.1 & 50.7 & 1.49s & 48.38M & 263\\
%			+Non-Local~\cite{wang2018non-local} & 38.9  & 23.1 & 42.4 & 50.3 & 1.60s & 49.95M & 265\\
%			%+ASPP~\cite{chen2018deeplab:} & 38.8 & 59.9 & 41.9 & 23.1 & 42.2 & 50.3\\
%			+DenseASPP~\cite{yang2018denseaspp} & 39.0 & 23.2 & 42.3 & 50.1 & 1.89s& 60.05M & 275\\
%			\hline
%		\end{tabular}
%	\end{center}
%	\caption{Performance of context module. Time means the training time of each iteration.}
%	\label{Table7}	
%\end{table}	
\textbf{Ablation studies for CRB.} Experiment results related with attention settings of Channel Refinement Block are presented in Table~\ref{Table6}. We experiment with several settings to get the channel weights. The first setting is applying Global Average Pooling (GAP) to the low-level feature layer. The second setting is applying GAP to the high-level feature layer. In the third setting, we downsample the low-level feature, add it to the high-level feature and finally apply GAP to the addition result. In the fourth setting, we concatenate the downsampled low-level feature with the high-level feature and finally apply GAP after dimension reduction. 
When using the second setting, it is equivalent to applying SE Block~\cite{hu2019squeeze-and-excitation} to the high-level feature layer, which is 1.2 AP higher than the baseline. When we adopt the first setting, an improvement of 1.9 AP can be achieved. The third and fourth settings are 1.7 AP and 1.7 AP higher than the baseline, almost equal to the first setting, but introduce more parameters. The results show that our proposed CRB considers multi-scale information from adjacent layers, which is helpful to guide the fusion of different scale features.

%\textbf{Ablation studies for context module.} Many object detection frameworks prove the importance of context information. %AugFPN~\cite{guo2019augfpn:} adds a ratio-invariant supplement composed of adaptive pooling at the highest level of FPN. ACFPN~\cite{Cao2020ACFPN} adds DenseASPP~\cite{yang2018denseaspp} to the highest level of FPN to increase the model receptive field. ThunderNet~\cite{qin2019thundernet:} merges multi-scale information on the top-level features to expand the receptive field. 
%As shown in Table~\ref{Table7}, PPM~\cite{zhao2017pyramid}, Non-local~\cite{wang2018non-local} and DenseASPP~\cite{yang2018denseaspp} are used in our ablation experiments. The results show that PPM obtains the best result of 39.1 AP, while other methods have similar performance. Finally, we choose PPM as our contextual head considering its faster speed and lower parameters.
\begin{table}
	\begin{center}
		\tiny
		\begin{tabular}{l| c c c c c c c }
			\hline
			Method & $AP$ & $AP_{S}$ & $AP_{M}$ & $AP_{L}$ & Time & Params & Flops(G) \\
			\hline\hline
			FPN &37.2 & 21.2 & 40.9& 48.1 & 1.14s & 41.53M  & 207\\
			DRFPN w/o PPM & 38.3  & 21.5 & 41.7 & 50.0 & 1.37s & 47.79M & 263\\
			+PPM~\cite{zhao2017pyramid} & 39.1  & 22.9 & 43.1 & 50.7 & 1.49s & 48.38M & 263\\
			+Non-Local~\cite{wang2018non-local} & 38.9  & 23.1 & 42.4 & 50.3 & 1.60s & 49.95M & 265\\
			%+ASPP~\cite{chen2018deeplab:} & 38.8 & 59.9 & 41.9 & 23.1 & 42.2 & 50.3\\
			+DenseASPP~\cite{yang2018denseaspp} & 39.0 & 23.2 & 42.3 & 50.1 & 1.89s& 60.05M & 275\\
			\hline
		\end{tabular}
	\end{center}
	\caption{Performance of context module. Time means the training time of each iteration.}
	\label{Table7}
\end{table}
	
\textbf{Ablation studies for context module.} Many object detection frameworks prove the importance of context information. %AugFPN~\cite{guo2019augfpn:} adds a ratio-invariant supplement composed of adaptive pooling at the highest level of FPN. ACFPN~\cite{Cao2020ACFPN} adds DenseASPP~\cite{yang2018denseaspp} to the highest level of FPN to increase the model receptive field. ThunderNet~\cite{qin2019thundernet:} merges multi-scale information on the top-level features to expand the receptive field. 
As shown in Table~\ref{Table7}, PPM~\cite{zhao2017pyramid}, Non-local~\cite{wang2018non-local} and DenseASPP~\cite{yang2018denseaspp} are used in our ablation experiments. The results show that PPM obtains the best result of 39.1 AP, while other methods have similar performance. Finally, we choose PPM as our contextual head considering its faster speed and lower parameters.

\subsection{Runtime Analysis}
We also measure the training and inference time when replacing FPN with DRFPN. As mentioned in the previous section, we set the batch size of each GPU to 8, and the total batch size to 24 for training. The training time of Faster-RCNN with ResNet50-DRFPN is about 1.37s per iteration and that of Faster-RCNN with ResNet50-FPN is 1.14s per iteration on COCO dataset. when using one RTX TITAN GPU, for an image with a shorter size of 800 pixels, the inference time of DRFPN is 12.4 ms and the inference time of FPN is 12.2 ms. The results show that DRFPN achieves better performance with almost the same inference time.

%------------------------------------------------------------------------

\section{Conclusion}

In this paper, we analyze the design defect of the feature merging of adjacent layers in FPN from pixel level and feature map level. For the existing problems, we design two modules: SRB and CRB, which respectively optimize the problems of inaccurate spatial sampling and inaccurate channel fusion. Then we demonstrate their efficacy by designing a new parameter-free feature pyramid network named DRFPN and report extensive experimental analysis showing that it achieves state-of-the-art performance.

{\small
	\bibliographystyle{ieee_fullname}
	\bibliography{egbib}
}

\end{document}